\documentclass[10pt,twocolumn,letterpaper]{article}

\usepackage{wacv}              

\usepackage{tabularx}
\usepackage{xspace}
\usepackage{multirow}
\usepackage{enumitem}
\usepackage{ragged2e}

\usepackage{graphicx}
\usepackage{amsmath}
\usepackage{amssymb}
\usepackage{booktabs}

\usepackage[pagebackref,breaklinks,colorlinks]{hyperref}

\usepackage[capitalize]{cleveref}
\crefname{section}{Sec.}{Secs.}
\Crefname{section}{Section}{Sections}
\Crefname{table}{Table}{Tables}
\crefname{table}{Tab.}{Tabs.}

\def\wacvPaperID{*****} 
\def\confName{WACV}
\def\confYear{2025}

\begin{document}

\title{Outcome-Guided Distillation: A Teacher-Student Framework to Advance VLM Reasoning in Autonomous Driving} 

\author{Zeyu Dong\\
Stony Brook University, USA\\
{\tt\small zeyu.dong@stonybrook.edu}
\and
Yimin Zhu\\
Stony Brook University, USA\\
{\tt\small yimzhu@cs.stonybrook.edu}
\and
Yu Wu\\
Rutgers University, USA\\
{\tt\small yu.wu@rutgers.edu}
\and
Yu Sun\\
Sunrise Technology Inc., USA\\
{\tt\small sunrisetechnology001@gmail.com}
}

\maketitle

\begin{abstract}
End-to-end (E2E) autonomous driving aims to learn a direct mapping from sensor observations to driving actions. However, these E2E models often act as black boxes and struggle with complex scenarios. To address this, recent works incorporate Vision-Language Models (VLMs) to provide explicit reasoning, enhancing both interpretability and driving robustness. These approaches, however, typically rely on pre-generated annotations, which can suffer from potentially flawed labels and require costly human labor. In this work, we propose a new framework that integrates structured reasoning and geometric precision through a teacher–student architecture. Our teacher model employs novel ``reflective reasoning,'' refining its reasoning under the supervision of a ground-truth action. This approach enhances zero-shot generalization without intermediate labels. A student model then learns the reasoning capability through distillation. To ensure geometric precision, we also design a separate waypoint decoder that translates the student's textual reasoning into continuous trajectories. Our solution integrates two goals: model interpretability and robust driving performance, using reasoning to explicitly guide prediction. Evaluated on Waymo benchmarks, our framework outperforms reasoning-based baselines in waypoint accuracy and efficiency. Notably, our experiments show that adding reasoning improves performance by nearly 24\% compared to an identical non-reasoning model, advancing a path toward deployable, interpretable driving systems. Our work advances reasoning-driven autonomous driving toward interpretable and deployable systems.
\end{abstract}

\section{Introduction}
End-to-end (E2E) autonomous driving has emerged as a compelling data-driven paradigm, aiming to learn a direct mapping from raw sensor inputs to vehicle control commands. This approach has significant contrasts with traditional modular designs that decompose the driving task into discrete, individually trained components, such as perception, prediction, and planning. Training neural networks in an end-to-end manner helps the systems avoid the error accumulation that affects cascaded modules and reduces the reliance on expensive, intermediate labels such as ground-truth object detections or semantic maps~\cite{bojarskiEnd2016,chenEndtoend2023}.
 
The evolution of E2E driving has been marked by the increasing sophistication of its underlying models. Early successes were built on pre-trained vision backbones, including Vision Transformers (ViT) or Bird's-Eye-View (BEV) encoders, that excelled at extracting features from complex visual scenes~\cite{shaoSafetyEnhanced2022,shaoReasonNet2023,huPlanningoriented2023}. Recently, there has been a growing trend to leverage the powerful capabilities of Vision-Language Models (VLMs) for enhancing semantic comprehension in the field. Early implementations employed  VLMs for tasks such as scene analysis or typical obstacle detection~\cite{xuDriveGPT42023, maoGPTDriver2023a}. Recent efforts move beyond simple perception and behavior cloning and leverage the Chain-of-Thought (CoT) reasoning for advanced cognitive driving tasks~\cite{hwangEMMA2024,wangDriveCoT2024,maoGPTDriver2023a}.

The integration of VLM-based reasoning represents a significant progress for two primary reasons. First, it addresses the critical challenge of interpretability. By generating textual analysis for its actions, the VLM clarifies the opaque nature of traditional E2E models, providing a clear, human-readable justification for its decisions. This interpretability is crucial for debugging, validation in safety-related events, and long-tail edge cases.  Once stakeholders understand the self-driving behavior, transparent reasoning traces build social trust in autonomous systems. Second, VLM reasoning significantly enhances driving reliability. By forcing the model to generate a step-by-step logical chain, we are leveraging its world knowledge by compelling it to perform robust and causal thinking. This reasoning mechanism serves as an effective regularizer, compelling the model to align its visual comprehension with human-like causal inference, such as ``pedestrian crossing, full stop, and wait,'' rather than merely acquiring the feature and action coincidences typically observed in behavior cloning.  This structured, multi-step thinking is critical for decomposing complex, long-tail scenarios into manageable steps, promising a new level of robustness that a direct pixel-to-action mapping struggles to achieve.

One significant challenge in applying VLMs lies in harnessing their reasoning capability. The predominant approach for training reasoning-driven VLMs requires a dataset with detailed CoT explanations that encompass scene understanding, prediction, and action planning. The data is typically generated either by using a suite of existing modular models (e.g., for segmentation and object detection) or through expensive, manual human labeling~\cite{wangDriveCoT2024,qianNuScenesQA2024}. This methodology is less scalable. In addition, the quality of the learned reasoning is fundamentally limited by the accuracy of the external models or the consistency of human annotators. Furthermore, while some studies use massive, zero-shot VLMs to avoid fine-tuning, their reasoning can be unreliable or even misleading. A central challenge, therefore, remains: how can we guide a VLM to produce high-quality, task-relevant reasoning without relying on intermediate supervision?

The second challenge lies in the VLM's inherent difficulty with generating precise, continuous numerical outputs, such as waypoints or control values. Many practices represent real numbers as text tokens~\cite{maoGPTDriver2023a}. This approach lacks intrinsic arithmetic computation capability, as it operates on statistical token patterns rather than symbolic numerical reasoning, ignoring the magnitude and precision difference among numbers. A quick example is that some LLMs struggle to distinguish between 9.11 and 9.9~\cite{junco_every_2025,schnabel_stop_2024,korzhov_stop_2023}. As a result, it discards the critical geometric and spatial relationships between coordinates.

The third challenge is that fine-tuning VLM could result in ``catastrophic forgetting.'' The conventional approach is to fully fine-tune the entire VLM model, including the vision encoder. While it specializes in modeling common driving scenarios, it is hindered by its broad, pre-trained world knowledge. Consequently, it weakens the generalization capability of VLMs for handling long-tailed driving scenarios.

To address these limitations, we propose a framework that integrates structured reasoning and geometric precision. Instead of relying on costly manual annotation, we introduce an outcome-guided "reflective reasoning" paradigm. By decoupling textual reasoning from numerical trajectory prediction, our system maintains interpretability while achieving high spatial accuracy. This approach allows us to leverage the world knowledge of large VLMs while maintaining enough efficiency for real-time deployment using a distilled student architecture.

Our main contributions are summarized as follows:
\begin{enumerate}
\item We introduce ``reflective reasoning,'' a novel teacher-student paradigm that resolves the challenge in joining two driving objectives: improving driving performance and enhancing interpretability with the innovative staged driving model. The model reveals intermediate reasoning and employs rational text to derive a high-confidence driving path. Utilizing inference with reasoning text results in a $24 \%$ improvement compared to inference without it.
\item We propose a decoupled architecture that uses a specialized, lightweight VLM as a waypoint decoder, enabling precise and stable numerical trajectory regression from textual reasoning.
\item We demonstrate that using a frozen vision encoder for the waypoint prediction model preserves essential world knowledge, avoiding catastrophic forgetting and leading to superior generalization and overall performance.
\end{enumerate}

\section{Related Work}
\paragraph{E2E Autonomous Driving}
E2E autonomous driving generates driving actions directly from raw sensor data~\cite{chenEndtoend2023}. Early work, such as PilotNet~\cite{bojarskiEnd2016} and Conditional Imitation Learning (CIL), demonstrated feasibility but suffered limited generalization~\cite{codevilla2018end}. Follow‑ups improved closed‑loop robustness via ``teacher'' agents~\cite{chen2020learning, chen2021learning}, image backbone~\cite{sonataEndtoEnd2023, khanumEndtoEnd2020}, transformer based fusion~\cite{chittaTransFuser2022,shaoSafetyEnhanced2022}, BEV image encoders~\cite{shaoReasonNet2023,chittaNEAT2021,huPlanningoriented2023}, and fine-tuning VLMs~\cite{shaoLMDrive2023}. 
A shared challenge of these approaches is the requirement for full model fine-tuning, which is both data- and computation-intensive. It also introduces a significant risk of overfitting to the training data.
In contrast, our approach preserves pretrained visual knowledge by freezing the vision encoder and focuses learning on reasoning and waypoint prediction, enabling better generalization with limited data.

\paragraph{Reasoning}
Recent advances in LLMs show the importance of inference time reasoning for robust decision making~\cite{wei2022chain, wang2022self, yao2022react, guan2024deliberative, zaremba2025trading}. In the driving domain, models such as DriveGPT4 and DriveCoT~\cite{xuDriveGPT42023,wangDriveCoT2024} generate CoT rationales, while RAG‑Driver~\cite{yuanRAGDriver2024} enhances reliability and context awareness with rationales in external knowledge. 
These insights motivate driving agents that explicitly allocate compute to “think” before committing to a trajectory, thereby narrowing the gap between black-box end-to-end control and deliberative planners. 
However, these methods rely on human-labeled reasoning or text datasets for supervision. Our ``reflective reasoning'' paradigm mitigates this dependency by allowing a teacher model to refine reasoning using ground-truth outcomes, then distilling it to a student model that reasons zero-shot without external labels.

\paragraph{VLM in Autonomous Driving} 
VLMs such as Flamingo~\cite{alayrac2022flamingo}, BLIP~\cite{li2022blip, li2023blip} and LLaVA~\cite{liu2024improved, liu2023visual} bridge visual and linguistic representations effectively. 
Autonomous driving VLMs broadly follow two paradigms: dual‑system approaches \cite{zhangADH2024,tianDriveVLM2024,dongGeneralizing2024} that use VLM to generate low‑frequency waypoints or high‑level commands to guide an E2E policy, and single‑system approaches \cite{maoGPTDriver2023a, hwangEMMA2024, zhang2024feedback} that reformulate trajectory prediction as text generation and interpret with CoT prompting. Text-centric pipelines suffer from inherent difficulty with precise and continuous numerical output generation. Our decoupled design addresses the numerical precision issue by utilizing a lightweight VLM specifically trained to interpret reasoning text and predict accurate waypoints.

\section{Methodology}\label{sec:c5-methodology}
We formulate E2E autonomous driving as a short-horizon trajectory planning task. At each timestep $T$, the agent receives a set of multi-view camera observations
\[
\mathcal{I}_T = \{ I_T^{(1)}, \dots, I_T^{(N_c)} \},
\]
captured from $N_c$ synchronized onboard cameras. In addition, the agent is provided with (i) a high-level driving command $c_T$ (e.g., \texttt{go-left}, \texttt{go-straight}, \texttt{go-right}), and (ii) a state history
\[
S_{T-k:T} = \{ s_{T-k}, \dots, s_{T-1}, s_T \},
\]
where each state $s_t$ may include position, velocity, heading, and other kinematic features.

The objective is to predict a sequence of $H$ future waypoints
\[
\hat{W}_T = (\hat{w}_{T+1}, \dots, \hat{w}_{T+H}), \qquad
\hat{w}_{t} \in \mathbb{R}^2,
\]
represented as 2D coordinates in the ego-centric frame of the vehicle. The learning problem is therefore to model the mapping
\[
f_{\theta} : (\mathcal{I}_T, c_T, S_{T-k:T}) \;\longrightarrow\; \hat{W}_T.
\]
Our objective is to ensure that the predicted trajectory is geometrically precise and meets the safety and comfort standards required for real-world driving. 

Our proposed framework differs from traditional modular pipelines and aims to create an efficient, end-to-end driving agent capable of reasoning and suitable for real-world, closed-loop deployment. Figure~\ref{fig:arch-distill} and~\ref{fig:arch-student} shows the model architecture that is built on three core principles: (i) a novel ``reflective reasoning'' paradigm to train a VLM for robust, zero-shot generalization without relying on intermediate labels; (ii) a teacher-student distillation process to transfer and compress the large model's capabilities into a light-weight, compact model for real-time inference; and (iii) a decoupled architecture that separates high-level textual reasoning from precise numerical waypoint prediction. 

\subsection{Enhancing Zero-Shot Generalization via Reflective Reasoning}\label{sec:reflect}

\begin{figure*}[t]
\centering
\includegraphics[width=\linewidth]{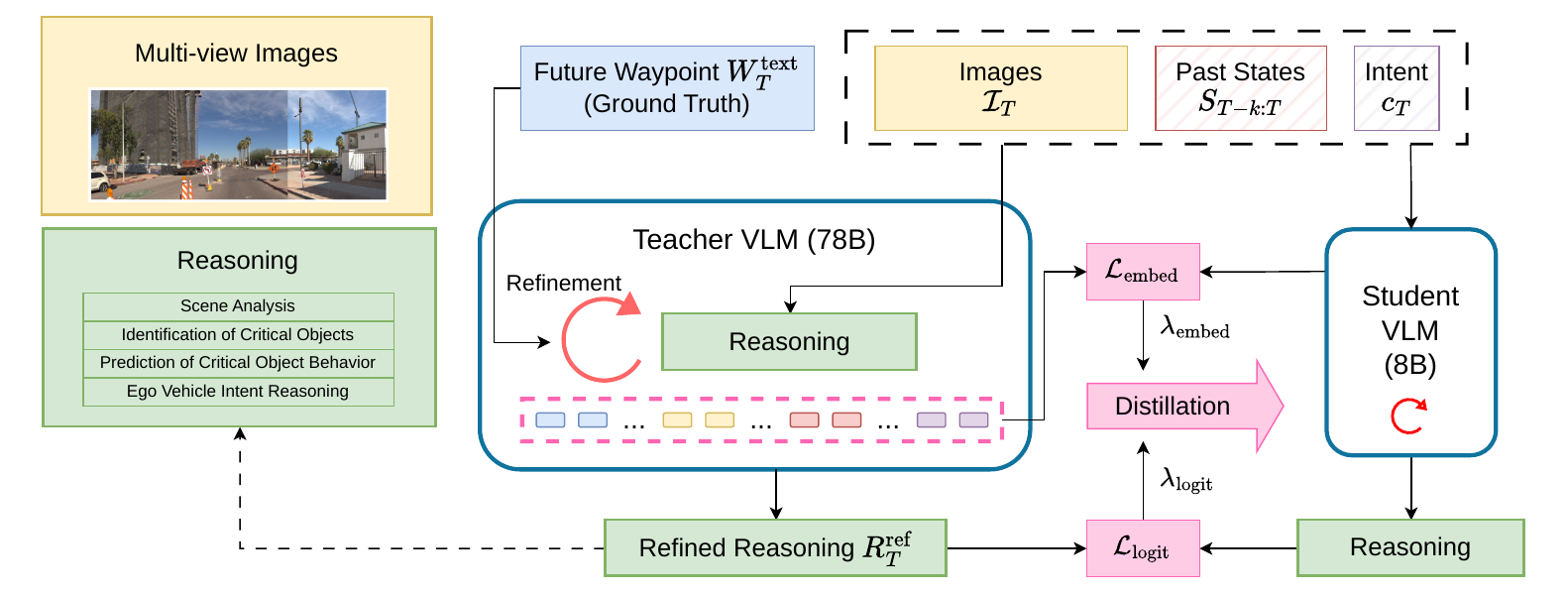} 
\caption{The distillation of the teacher and student model distillation process; left: an illustration of the input image and the output reasoning; middle: the teacher using the ground truth waypoint to refine its reasoning based on the input; right: the student model learning the refined reasoning through distillation.}
\label{fig:arch-distill}
\end{figure*}

We introduce a new training paradigm called \textbf{reflective reasoning}, implemented in the teacher VLM. As shown in the middle part of the Figure~\ref{fig:arch-distill}, instead of one-directional and single-round reasoning, we trigger the bidirectional and multiple-round reasoning with the guidance from the groundtruth waypoint by following those steps for each training sample:

\paragraph{Input sampling.} As shown on the top of the Figure~\ref{fig:arch-distill}, we first select one training example consisting of visual observations, high-level driving command, and state history: $(\mathcal{I}_T, c_T, S_{T-k:T})$, along with the expert future waypoints $W_T$.

\paragraph{Waypoint verbalization.} Convert the numerical waypoint sequence $W_T$ into a compact natural-language description $W_T^{\text{text}}$ (e.g., ``move forward 8\,m, then curve slightly right and stop at the intersection'').

\paragraph{Reasoning generation.} Illustrated by the Teacher VLM in Figure~\ref{fig:arch-distill}, we use multi-round conversations with the teacher model to refine the reasoning under the guidance of the ground truth waypoint description.
\begin{enumerate}[label=\alph*)]
\item bottom-up reasoning: provide the VLM with the input features and prompt it to generate a forward reasoning.
\item top-down reasoning: provide the VLM additionally with the ground truth waypoint description, and prompt it to generate a step-by-step, causal backward reasoning that justifies why these waypoints are appropriate for the observed scene.
\item reflective summary: collect the bidirectional reasonings and prompt the VLM to summarize and refine them into a final reflective reasoning $R_T^{\text {ref}}$.
\end{enumerate}

\paragraph{Dataset construction.} Store the resulting tuple
\[
(\mathcal{I}_T, c_T, S_{T-k:T}, R_T^{\text {ref }}),
\]
which forms one training element for later distillation.

Besides the conventional one-directional CoT, where the model simply imitates a pre-written explanation, reflective reasoning forces the VLM to \emph{derive} the missing reasoning that logically connects the scene to the expert action (ground truth). This introspective, outcome-conditioned generation encourages the VLM to internalize a more generalizable causal structure. Therefore, this approach improves zero-shot reasoning in unseen scenarios without requiring intermediate labels, such as object boxes, affordance maps, or human-written rationales. The key differences are summarized in Table~\ref{tab:reasoning-comparison}.


\newcolumntype{L}{>{\RaggedRight\arraybackslash}X}

\begin{table*}[t]
\centering
\small 
\begin{tabularx}{\textwidth}{@{} l L L @{}}
\toprule
\textbf{Property} & \textbf{Standard CoT} & \textbf{Reflective Reasoning (Ours)} \\
\midrule
\textbf{Reasoning Process} & Single-pass, forward generation (Scene $\rightarrow$ Reasoning $\rightarrow$ Action). & Multi-pass: forward (bottom-up), backward (top-down), and refinement (reflection). \\
\addlinespace[10pt]
\textbf{Reasoning Quality} & Prone to hallucination, topic drift, or causal errors. & Reasoning is anchored by the ground-truth expert outcome. \\
\addlinespace[10pt]
\textbf{Supervision Source} & Relies on pre-generated, static CoT labels (human or model-based). & No CoT labels needed. Uses expert waypoints, which are standard dataset components. \\
\bottomrule
\end{tabularx}
\vspace{.1in}
\caption{Key differences between standard CoT and our proposed reflective reasoning paradigm. Our method is constrained by the expert outcome, forcing it to generate causally-grounded logic rather than unconstrained, plausible-sounding text.}\label{tab:reasoning-comparison}
\end{table*}

\subsection{Teacher-Student Distillation for Efficient Deployment}
After being extracted from a larger teacher model, the reflective reasonings are used for knowledge distillation. The purpose of the distillation stage is to transfer the reasoning capability of the large teacher VLM into a compact student model that can be deployed on the edge computer to perform in-device inference in real-time. More importantly, the student is challenged to generate the reasoning without any preloaded ground truth waypoints to make the system practical in deployment. At this stage, the student is trained \emph{only} to generate structured reasoning; the prediction of numerical waypoints is deferred to the downstream RealNum-Decoder module introduced in the next subsection.

As shown on the right side of Figure~\ref{fig:arch-distill}, the student is provided with the dataset of distillation tuples
\[
\mathcal{D}_{\text{distill}} = \{(\mathcal{I}_T, c_T, S_{T-k:T}, R_T^{\text {ref }})\},
\]
The student model is trained to reproduce the teacher's reasoning output in an autoregressive manner, using the following two complementary distillation signals:

\begin{enumerate}[leftmargin=1.5em]
\item \textbf{Token-level distillation.}
For the logits of each reasoning token $r_i$, the student matches the teacher's next-token distribution via cross-entropy loss:
\[
\mathcal{L}_{\text{logit}} = -\log\frac{\exp{(r_i)}}{\sum_k^{|D|}\exp{(r_k)}},
\]
where $|D|$ is the size of the token dictionary. This explicitly transfers the teacher's linguistic reasoning capability at the token level.

\item \textbf{Embedding-level distillation.}
Before the teacher's language model head projects hidden states to logits, we extract the hidden embedding $h_i^{\mathrm{teacher}}$. The student is trained to match this internal representation with \(L_2\) loss:
\[
\mathcal{L}_{\text{embed}} = \left\|h_i^{\mathrm{student}} - h_i^{\mathrm{teacher}}\right\|_2^2.
\]
This encourages the student to reproduce the teacher's \emph{latent reasoning structure}, rather than mimicking the teacher's surface text pattern.
\end{enumerate}

The full distillation objective is:
\[
\mathcal{L}_{\text{distill}} =
\lambda_{\mathrm{logit}} \mathcal{L}_{\mathrm{logit}}
+ \lambda_{\mathrm{embed}} \mathcal{L}_{\mathrm{embed}},
\]
where $\lambda_{\mathrm{logit}}$ and  $\lambda_{\mathrm{embed}}$ are hyperparameters to adjust the weight of two losses. After training, the student model will generate the same structured reasoning as the teacher, which will be consumed by the RealNum-Decoder for numerical waypoint regression. 

\subsection{RealNum-Decoder for Numerical Waypoint Prediction}\label{sec:num-traj}

\begin{figure}[t]
\centering
\includegraphics[width=0.6\linewidth]{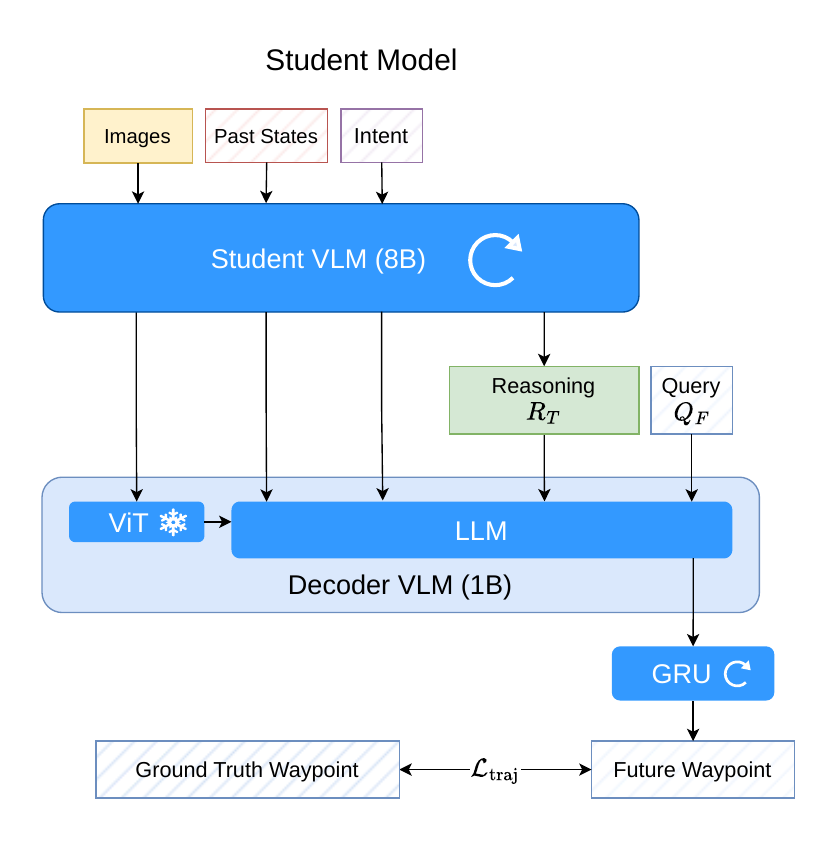} 
\caption{The architecture of the RealNum-Decoder; }
\label{fig:arch-student}
\end{figure}


 Figure~\ref{fig:arch-student} displays the proposed  RealNum-Decoder that adopts a decoupled design: while the student VLM in the previous subsection learns to generate structured reasoning, \emph{RealNum-Decoder} converts that structured reasoning into precise, continuous waypoints. The two different models are trained with separate objective functions. This design successfully preserves the linguistic generalization of the student while achieving geometric accuracy for waypoint prediction.  The detail about this pipeline is shown as follows:

\paragraph{(a) Reasoning Encoder (Student VLM).}
After the distillation stage, we utilize the distilled student model as a reasoning encoder, which takes the driving inputs and generates the reasoning text. Concretely, given
\[
(\mathcal{I}_T, c_T, S_{T-k:T}),
\]
the encoder autoregressively generates a reasoning sequence $R_T=\{r_1,\dots,r_L\}$, where $L$ is the length of reasoning and $r_i$s are the tokens. Then, together with a trainable Future Way Point Query $Q_\text{F}$, and other input features, the reasonings are sent to downstream trajectory regression via a Decoder VLM and a subsequent Gated Recurrent Unit (GRU).

\paragraph{(b) RealNum-Decoder (Decoder VLM + GRU).}
To generate the future trajectory, we first feed the ditilled reasoning $R_T$, the trainable Future Waypoint Query $Q_{\text{F}}$, and all scene-related inputs $(\mathcal{I}_T, c_T, S_{T-k:T})$ into the Decoder VLM $g_{\text{D}}$, which integrates the reasoning with the visual, semantic, and historical features to produce a unified latent representation. Notice that the image $\mathcal{I}_T$ is sent to a ViT in Decoder VLM while the other components are sent to a trainable LLM. The ViT is pre-trained and also frozen during training to keep its capacity for generalization. This outcome representation is then passed into a GRU module $g_{\text{GRU}}$, which models temporal dependencies and refines the fused features into a sequence of future waypoint predictions. The final output $\hat{W}_T \in \mathbb{R}^{H \times 2}$ corresponds to $H$ predicted 2D waypoints, forming the vehicle's future trajectory over a horizon of $H$ within the frame:

\[
\hat{W}_T = g_{\text{GRU}}(g_{\text{D}}(R_T, Q_\text{F},  \mathcal{I}_T, c_T, S_{T-k:T}))\in \mathbb{R}^{H\times 2}.
\]
\paragraph{Training Objective.}
Instead of using a $L_2$ (MSE) or $L_1$ loss function, we adopt a tolerance-weighted worst-axis displacement objective that emphasizes safety-critical deviations along lateral and longitudinal axes. For waypoint $t$:
\[
\mathcal{L}_0(w_t, \hat{w}_t) = \max \left\{ \frac{\Delta_{\text{lat},t}}{\tau_{\text{lat},t}}, \frac{\Delta_{\text{lng},t}}{\tau_{\text{lng},t}} \right\},
\]
where $\Delta_{\text{lat},t}=|w_{\text{lat},t}-\hat{w}_{\text{lat},t}|$ and $\Delta_{\text{lng},t}=|w_{\text{lng},t}-\hat{w}_{\text{lng},t}|$ are the lateral and longitudinal errors for the $t$-th waypoint, and $\tau_{\text{lat},t}, \tau_{\text{lng},t}$ are pre-defined tolerance thresholds~\cite{Visionbased2025}. The overall trajectory loss averages across the horizon:
\[
\mathcal{L}_{\text{traj}}(W_T, \hat{W}_T) \;=\; \frac{1}{H} \sum_{t=1}^{H} \mathcal{L}_0(w_t, \hat{w}_t).
\]
This objective can be interpreted as an $L_\infty$ penalty for the weighted errors along each axis at every timestep, followed by an $L_1$-style aggregation across timesteps. This approach combines the robustness against the worst-case deviations with stable optimization~\cite{qifakeRobust2005}. In practice, we normalize coordinates to the frame and apply per-dataset scaling to $\tau_{\cdot,t}$ for consistent gradient magnitudes.

\paragraph{Discussion.}
Decoupling offers two advantages: (i) the Reasoning Encoder (student VLM)'s language competence and zero-shot reasoning remain intact. It is trainable only in the distillation stage and only doing inference in the RealNum-Decoder training stage; (ii) the RealNum-Decoder focuses entirely on the geometry using a loss function tailored for driving safety. This division of labor allows each component to perform its specialized task optimally, preserving the VLM's powerful reasoning capabilities while achieving accurate and stable numerical prediction. At the inference stage, the end-to-end path is:
\[
(\mathcal{I}_T, c_T, S_{T-k:T}) \xrightarrow{\text{Student}} R_T
\xrightarrow{\text{RealNum-Decoder}} \hat{W}_T.
\]
This preserves interpretability via $R_T$ while delivering precise, real-time waypoint regression via RealNum-Decoder.




\section{Experiments}\label{sec:experiments}
To validate our proposed framework, we conduct a comprehensive set of experiments on the Waymo E2E Driving dataset~\cite{waymo2025e2edriving}. Our evaluation is designed to rigorously test our core contributions. Specifically, we aim to:
\begin{itemize}[leftmargin=*,itemsep=0pt,topsep=3pt,parsep=3pt]
\item Validate our outcome-guided reflective reasoning paradigm and analyze the efficiency of our teacher-student distillation process.
\item Evaluate the use of a decoupled waypoint decoder for precise numerical prediction.
\item Conduct ablation studies on the importance of using a frozen vision encoder and the value of reasoning itself.
\end{itemize}

\subsection{Experimental Setup}

\paragraph{Datasets}
We use the Waymo E2E Driving dataset, a large-scale, waypoint-prediction-focused dataset. We report two types of metrics: the official Rater Feedback Score (RFS), where higher is better, and standard motion-planning metrics (ADE, ADE@3s, ADE@5s), where lower is better.

\paragraph{Baselines and Model Variants}
We compare our proposed method against a series of baselines and model variants to isolate the impact of our contributions.
\begin{itemize}
\item \textbf{Our Method:} Consists of a distilled InternVL3-8B student model, trained on data from the teacher's reflective reasoning, and RealNum-Decoder with a frozen ViT for waypoint prediction.
\item \textbf{Teacher:} Consists of the full InternVL3-78B teacher model and RealNum-Decoder with the same configuration. The teacher model generates the reflective reasoning as introduced in Section~\ref{sec:reflect}. This serves as a practical upper bound for distillation.
\item \textbf{Finetuned ViT:} A variant of our approach, but the ViT in RealNum-Decoder is fully finetuned on the driving dataset instead of being kept frozen.
\item \textbf{Text-Based Waypoints:} A variant of our approach that replaces the decoupled waypoint decoder with a standard text-based approach, where waypoint coordinates are tokenized and generated auto-regressively.
\item \textbf{Standard CoT:} We use the 78B teacher model and the RealNum-Decoder with the same configuration of our method, but prompted to generate a standard CoT reasoning without the guidance of our reflective reasoning outcome-based paradigm.
\item \textbf{Direct Prediction:} An 8B model with a frozen ViT and the GRU waypoint decoder, but trained directly on \((\mathcal{I}_T, c_T, S_{T-k:T},W_T)\) pairs without any reasoning involves. This isolates the performance impact of the reasoning component itself.
\end{itemize}

\subsection{Quantitative Analysis}

\begin{table*}[htbp]
\centering
\resizebox{\textwidth}{!}{%
\begin{tabular}{l c c c c c c c c}
\toprule
\textbf{Model} & \textbf{Size} & \textbf{ViT State} & \textbf{Decoder} & \textbf{Reasoning} & \textbf{Waymo (RFS) $\uparrow$} & \textbf{ADE $\downarrow$} & \textbf{ADE@3s $\downarrow$} & \textbf{ADE@5s $\downarrow$} \\
\midrule
Direct Prediction & 8B & Frozen & Waypoint & None & 5.849 & 7.106 & 5.989 & 8.222 \\
Standard CoT & 78B+1B & Frozen & Waypoint & Standard & 6.536 & 3.289 & 2.599 & 3.979 \\
Text-Based Waypoints & 8B+1B & Frozen & Text-based & Reflective & 5.690 & 6.089 & 4.151 & 8.028 \\
Finetuned ViT & 8B+1B & Finetuned & Waypoint & Reflective & 6.554 & 4.338 & 3.632 & 5.044 \\
\textbf{Our Method} & \textbf{8B+1B} & \textbf{Frozen} & \textbf{Waypoint} & \textbf{Reflective} & \textbf{7.240} & \textbf{3.231} & \textbf{2.726} & \textbf{3.735} \\
\midrule
Teacher (Upper Bound) & 78B+1B & Frozen & Waypoint & Reflective & 7.639 & 3.197 & 2.944 & 3.451\\
\bottomrule
\end{tabular}%
}
\vspace{.1in}
\caption{Main results comparing our method against baselines on the Waymo E2E Driving dataset. Our 8B method outperforms all comparable baselines, validating our contributions in reflective reasoning, decoupled decoding, and using a frozen ViT. It also successfully closes the gap to the 78B Teacher.}\label{tab:main_results}
\end{table*}

Our main quantitative results are summarized in Table~\ref{tab:main_results}. The analysis validates our architectural choices, as our final 8B model achieves an RFS of 7.240, outperforming all other 8B baselines by a significant margin.

First, the results demonstrate that reasoning is essential for high-quality planning. The Direct Prediction baseline, which lacks a reasoning module, performs poorly with an RFS of 5.849 and an ADE of 7.106. This suggests that forcing the model to generate a logical, textual explanation acts as a powerful regularizer, leading to a much more robust and generalizable planning policy.

Furthermore, our architectural choices for handling VLM-based reasoning is critical. The Text-Based Waypoints baseline, which tokenizes coordinates, is the worst-performing model, scoring only 5.690 RFS. Its trajectory error is also high, at 6.089 ADE. This confirms our hypothesis that using LLMs for direct numerical regression is fundamentally flawed. Our decoupled waypoint decoder, by contrast, achieves a dramatically better ADE of 3.231.

We also validate our ``reflective reasoning'' paradigm by comparing the 78B Teacher against the 78B standard CoT model. Our outcome-guided method achieves an RFS of 7.639, substantially stronger than the standard CoT approach's 6.536. This indicates that unconstrained reasoning is less reliable. In contrast, our reflective reasoning paradigm produces reasoning logic that is far more aligned with the driving task.

Finally, the comparison to the Finetuned ViT baseline validates our decision to keep the encoder frozen. The finetuned model performs worse than our frozen-encoder model, scoring 6.554 on RFS and 4.338 on ADE. This strongly suggests that finetuning on a specific driving dataset causes catastrophic forgetting. As a result, the VLM's rich and pre-trained world knowledge is lost during the training, making the model less generalizable.

\subsection{Performance and Efficiency Analysis}
A primary goal of our work is to create an agent that is not only accurate but also practical for deployment. Therefore, it is necessary to distill the capabilities of the massive 78B teacher model into a compact and efficient student model. We analyze this trade-off in Table~\ref{tab:efficiency}. All efficiency-related metrics are benchmarked on two A6000 (48G) GPUs.

The results clearly justify the advantage of distillation. Our final system, with a total size of 9B parameters (8B student + 1B decoder), is approximately eight times smaller than the 78B Teacher. This size reduction translates to a massive gain in computational speed. The 78B Teacher VLM is slow, requiring 454.88 ms per generated token. In contrast, our 8B student VLM is nearly seven times faster, requiring only 68.30 ms per token. The specialized 1B waypoint decoder, which is not auto-regressive, has a fast and fixed inference latency of only 76.50 ms.

This component-level efficiency makes our system towarding to practical applications. Crucially, this efficiency gain comes at a minimal cost to performance. Our distilled system successfully retains the vast majority of the teacher's capabilities, scoring 7.240 RFS and 3.231 ADE, compared to the teacher's 7.639 RFS and 3.197 ADE. This demonstrates that our distillation approach is highly effective at transferring the complex reasoning and prediction capabilities of a massive VLM into a compact agent.

\begin{table}[htbp]
\centering
\resizebox{0.6\columnwidth}{!}{%
\begin{tabular}{l c c c}
\toprule
& \multirow{2}{*}{\textbf{Teacher}} & \multicolumn{2}{c}{\textbf{Our Method}} \\
\cmidrule(lr){3-4}
\textbf{Metric} & & \textbf{Student VLM} & \textbf{Decoder} \\
\midrule
\multicolumn{4}{l}{\textit{Efficiency}} \\
Model Size & 78B & 8B & 1B \\
Token Speed (ms/token) $\downarrow$ & 454.88 & 68.30 & N/A \\
Inference Latency (ms) $\downarrow$ & N/A & N/A & 76.50 \\
\midrule
\multicolumn{4}{l}{\textit{Performance}} \\
Waymo (RFS) $\uparrow$ & 7.639 & \multicolumn{2}{c}{7.240} \\
ADE (Overall) $\downarrow$ & 3.197 & \multicolumn{2}{c}{3.231} \\
ADE@3s $\downarrow$ & 2.944 & \multicolumn{2}{c}{2.726} \\
ADE@5s $\downarrow$ & 3.451 & \multicolumn{2}{c}{3.735} \\
\bottomrule
\end{tabular}%
}
\vspace{.1in}
\caption{Efficiency and Performance Trade-off. Our method is significantly faster than the 78B Teacher while retaining most of its performance. All latency tests run on two A6000 GPUs.}\label{tab:efficiency}
\end{table}

\begin{figure*}[t]
\centering
\includegraphics[width=\linewidth]{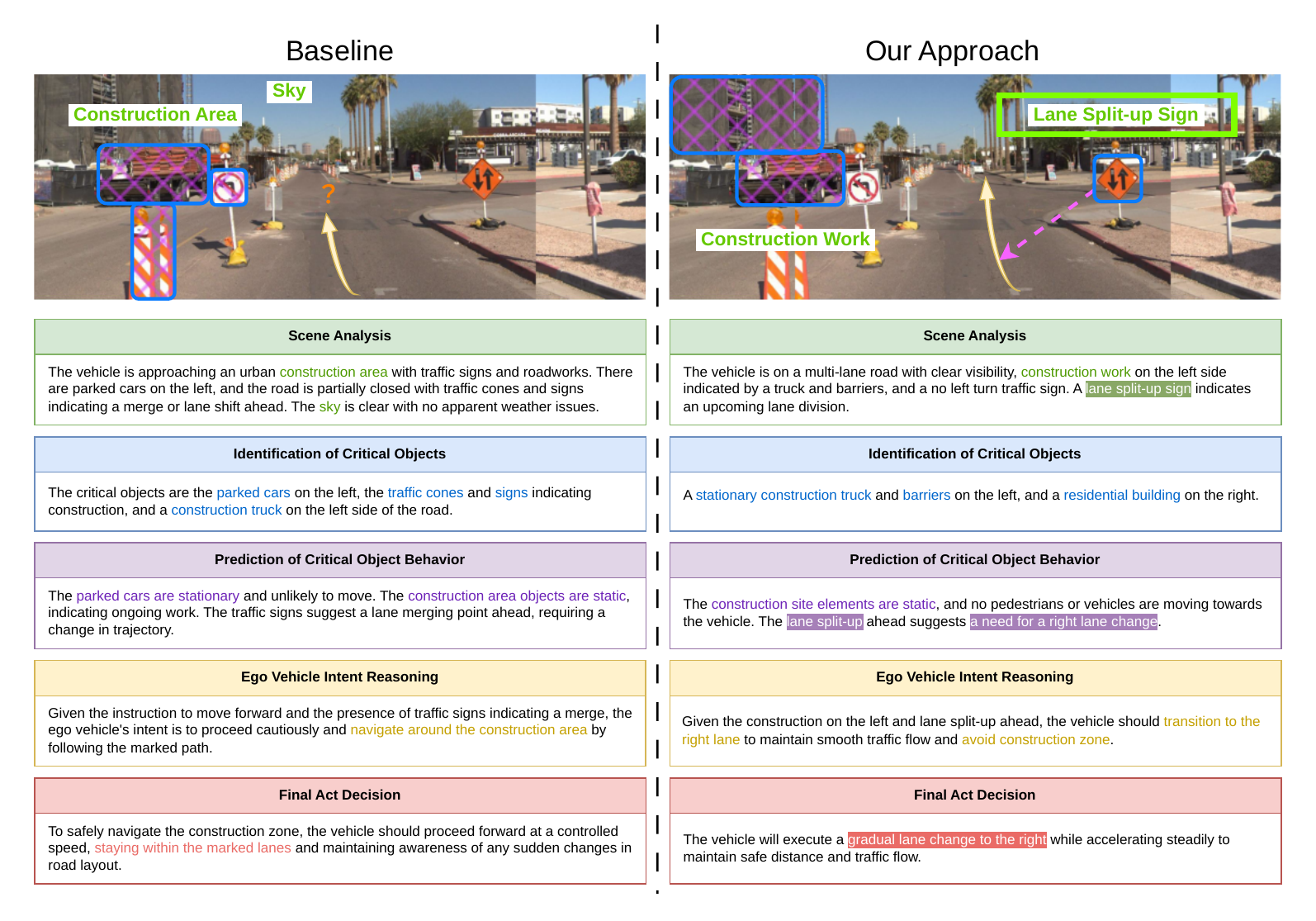}
\caption{Qualitative comparison of our reflective reasoning against the standard CoT baseline in a complex lane-split scenario. The baseline's low-quality CoT fails to identify the critical lane sign. Our model generates high-quality, causal logic that correctly links the lane sign to the required right-merge maneuver.}\label{fig:qualitative}
\label{fig:ana-com}
\end{figure*}

\subsection{Qualitative Results}

We provide qualitative visualizations of our model's outputs in Figure~\ref{fig:qualitative}. The figure illustrates a complex and multi-lane scenario. In this scenario, the left lane is closed and right lane splits into the two-way traffic because of the construction. Thus, the expert's ground-truth action is to merge right. The standard CoT baseline correctly identifies the general scene and important objects like other vehicles and the construction object. However, it fails to perform the correct causal reasoning. Its reasoning is unfocused and does not identify the single most critical object for this decision: the overhead lane split-up sign.

In contrast, our reflective reasoning paradigm demonstrates its core advantage. By being constrained by the expert outcome (merge right) during its training data generation, the teacher model is forced to find the specific visual evidence that justifies this action. It correctly ``reflects'' its reasoning process and identifies the lane split-up sign as the primary causal for the maneuver. Its reasoning is precise and high-quality, explicitly linking the lane sign to the necessary action of shifting to the right lane. This demonstrates how our outcome-guided process produces superior, causally correct reasoning that leads to a better driving performance.

\section{Conclusion}
In this paper, we presented a novel teacher-student framework designed to address three critical challenges in VLM-based autonomous driving: the reliance on costly CoT labels, poor numerical precision, and catastrophic forgetting from fine-tuning. We successfully addressed these challenges through a three-part solution. We introduced ``reflective reasoning,'' a label-efficient paradigm where a teacher model generates robust, causally-grounded logic by being constrained to justify a known expert outcome. This process, combined with a frozen vision encoder, preserves critical pre-trained knowledge and avoids catastrophic forgetting. Finally, our decoupled waypoint decoder architecture demonstrates how to successfully translate this textual reasoning into precise, geometrically stable trajectories, solving the VLM's inherent difficulty with numerical regression.

Our experiments on the Waymo E2E Driving dataset demonstrate the efficacy of our approach. Our approach significantly outperforms all comparable baselines in both safety (RFS) and accuracy (ADE). We validate that each of our components provides a significant benefit, and our final distilled model is efficient, maintaining the vast majority of the 78B teacher's performance while being faster at token generation. By creating a system that is interpretable, accurate, and computationally efficient, our work represents a significant step toward developing practical and deployable end-to-end driving agents.

\clearpage
\setcounter{page}{1}
{
\newpage
\centering
\Large
\vspace{0.5em}Supplementary Material \\
\vspace{1.0em}
}

\section{Training Setup and Details}

In this section, we provide a detailed breakdown of our experimental framework. We first describe the composition and preprocessing of the Waymo dataset used for our benchmarks. Next, we specify the exact training hyperparameters and hardware configurations for both the student VLM and the waypoint decoder. Finally, we present the training loss curves to demonstrate the stability and convergence of our two-stage training process.

\subsection{Dataset Details}\label{supp:dataset}

Our study is conducted using the Waymo Vision-based End-to-End Driving dataset, a large-scale benchmark deliberately curated to capture long-tail driving scenarios. The dataset covers diverse environments and rare events that occur with a frequency of less than 0.003\% in daily driving~\cite{waymo2025e2edriving}, such as navigating construction zones during public gatherings, avoiding fallen pedestrians, and handling unexpected freeway obstacles. These characteristics make the dataset a challenging and valuable benchmark for advancing generalizable, end-to-end autonomous driving capabilities, particularly for the task of future waypoint planning, which presents significantly greater complexity than classical perception tasks.

The dataset comprises 4,021 unique driving segments, with each segment capturing a continuous 20-second sequence sampled at 4\,Hz. This results in a total of 415,663 training samples. The task is to predict the vehicle's future trajectory based on its current sensor inputs and state history. As illustrated in Figure~\ref{fig:raw-input}, the inputs provided are:

\begin{figure}[ht]
\centering
\includegraphics[width=0.6\linewidth]{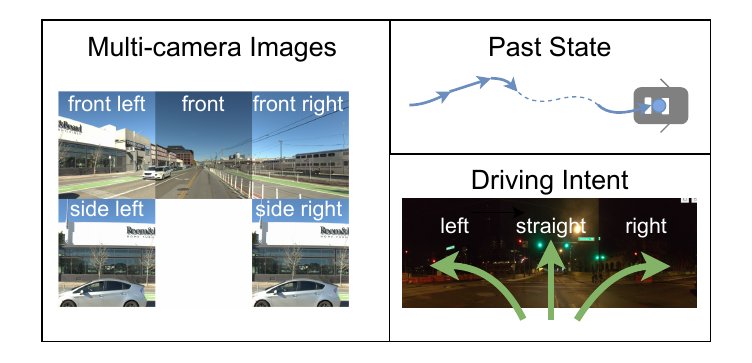}
\caption{Example of raw inputs provided in the Waymo dataset, including multi-view camera frames, state history, and high-level instruction.}
\label{fig:raw-input}
\end{figure}

\begin{itemize}
\item \textbf{Multi-camera Images ($\mathcal{I}$):} The vehicle is equipped with 8 cameras providing a 360-degree field of view. For computational efficiency, we utilize a subset of $N_c = 5$ cameras that provide comprehensive forward and side-facing coverage: \textit{front, front-left, front-right, side-left, and side-right}.
\item \textbf{Past State ($S_{\text{past}}$):} A sequence of the vehicle's kinematic state, including location, velocity, and acceleration over the past 4 seconds (sequence length of 16).
\item \textbf{Driving Instruction ($c$):} A high-level navigational command from the set $\{\texttt{go \allowbreak{}left},\ \allowbreak\texttt{go \allowbreak{}straight},\ \allowbreak\texttt{go \allowbreak{}right}\}$, which guides the vehicle's decision-making.
\end{itemize}

The model's objective is to output the future waypoints $\hat{W}$ for the next 5 seconds, corresponding to a prediction horizon of $H=20$.

\paragraph{Data Subsampling.} 
Due to the high computational cost associated with training large Vision-Language Models (VLMs) on video data, we apply a subsampling strategy to the training set. We utilize approximately 57\% of the total available training samples. This subset was selected to maintain the distribution of driving scenarios while fitting within our computational budget.

\subsection{Training Parameter Setup}
We implemented our framework using PyTorch and DeepSpeed to optimize training efficiency. To balance memory constraints with training speed, we utilized CPU offloading. All models were trained on a cluster node equipped with two NVIDIA A6000 (48GB) GPUs. We employed the Adam optimizer with a linear learning rate scheduler for all experiments.

The specific hyperparameters for the different model variants and baselines discussed in the experiments are detailed in Table~\ref{tab:training_params}. The evaluation metrics reported in the main paper are derived from the best-performing checkpoint for each model.

\begin{table}[htbp]
\centering
\caption{Training hyperparameters for the proposed architecture and baselines. All models use the Adam optimizer and a linear scheduler.}
\label{tab:training_params}
\resizebox{0.6\linewidth}{!}{%
\begin{tabular}{l c c c c}
\toprule
\textbf{Model Variant} & \textbf{BS} & \textbf{Epochs} & \textbf{LR} & \textbf{LoRA} \\
\midrule
\textbf{Student Distillation (8B)} & 128 & 2 & $5 \times 10^{-5}$ & Yes \\
\textbf{Waypoint Decoder (1B)} & 32 & 1 & $1 \times 10^{-5}$ & No \\
\midrule
\textit{Baselines} & & & & \\
Finetuned ViT (1B) & 128 & 1 & $5 \times 10^{-5}$ & No \\
Direct Prediction (1B) & 128 & 2 & $5 \times 10^{-5}$ & No \\
Text-Based Waypoints (1B) & 128 & 1 & $5 \times 10^{-5}$ & No \\
\bottomrule
\end{tabular}%
}
\end{table}

\subsection{Training Curves}
\begin{figure*}[htbp]
\centering
\begin{subfigure}[b]{0.32\linewidth}
\centering
\includegraphics[width=\linewidth]{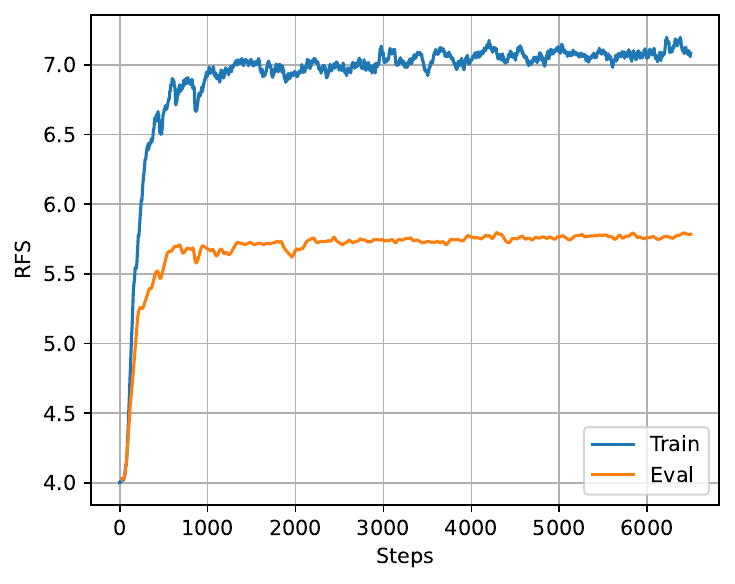}
\caption{Direct Prediction}
\label{fig:direct_pred}
\end{subfigure}
\hfill 
\begin{subfigure}[b]{0.32\linewidth}
\centering
\includegraphics[width=\linewidth]{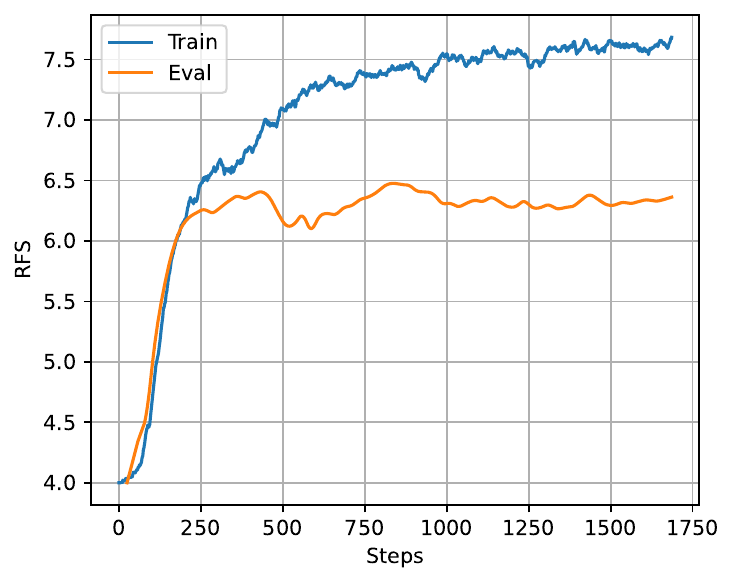}
\caption{Finetuned ViT}
\label{fig:finetuned_vit}
\end{subfigure}
\hfill 
\begin{subfigure}[b]{0.32\linewidth}
\centering
\includegraphics[width=\linewidth]{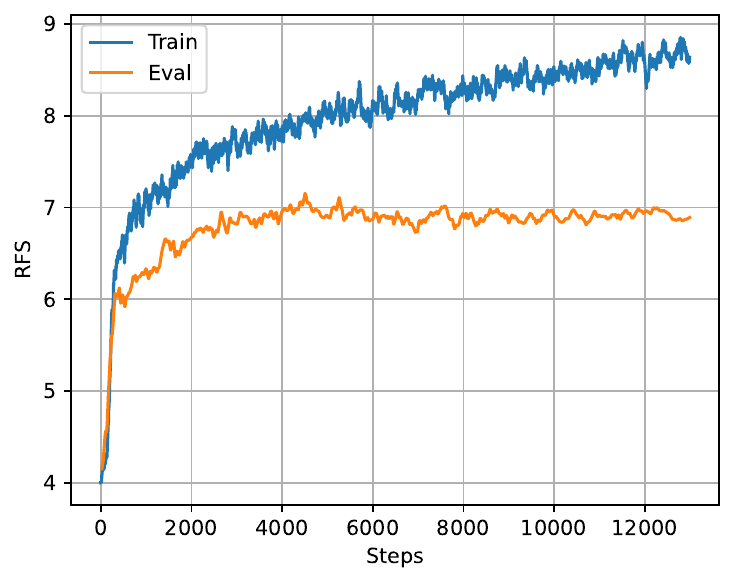}
\caption{Our Method}
\label{fig:our_method}
\end{subfigure}
\caption{Training loss curves across different methods.}\label{fig:training_curves}
\end{figure*}

To demonstrate the stability of our training process, we provide the training loss curves for the waypoint decoder in Figure~\ref{fig:training_curves}. The curves indicate that the models converge steadily without exhibiting signs of severe overfitting. Ensuring both the baseline method and our method are well-tuned for a fair comparison.

\section{RealNum-Decoder Details}
In this section, we provide details about the RealNum-Decoder and the motivation behind its training objective.

\subsection{Model Architecture}
As illustrated in Figure~\ref{fig:arch-student}, the waypoint prediction model is a compact Vision-Language Model (VLM) designed to interpret the structured reasoning process and, conditioned on that reasoning, generate precise future waypoints. The model processes four distinct input modalities:
\begin{itemize}
\item The multi-view camera observations ($\mathcal{I}$).
\item The vehicle's past kinematic states ($S_{\text{past}}$).
\item The high-level driving instruction ($c$).
\item The generated reflective reasoning content ($R^{\text{ref}}$).
\end{itemize}

We utilize the 1B parameter InternVL VLM as the architectural backbone. The visual inputs, driving instructions, and reasoning text are inherently aligned with the VLM's multi-modal input space. To incorporate the vehicle's kinematic history, which consists of continuous numerical values, we employ a specialized multi-layer 1D-CNN encoder. This encoder projects the sequence of past states into a single latent embedding, which is injected into the VLM as a special \texttt{[STATE]} token.

\begin{figure}[ht]
\centering
\includegraphics[width=0.7\linewidth]{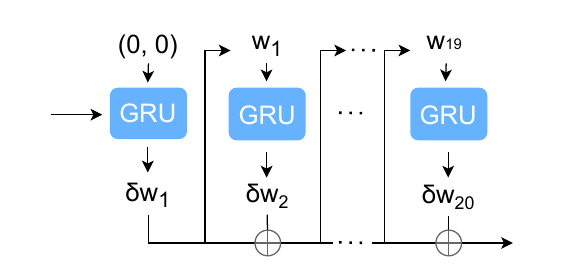}
\caption{Detailed architecture of the GRU-based waypoint decoder. The GRU takes the fused embedding from the VLM and autoregressively predicts offsets for the future trajectory.}
\label{fig:arch-gru}
\end{figure}

To synthesize these multimodal inputs for trajectory generation, a trainable \texttt{[ACTION]} token is appended to the end of the input sequence. The VLM processes the full context and produces a final output embedding corresponding to this action token. This embedding serves as the initial state for a downstream GRU module. Finally, as shown in Figure~\ref{fig:arch-gru}, the GRU operates autoregressively to decode the sequence of future waypoints $\hat{W}$.

\subsection{Training Objective}
In Section~\ref{sec:num-traj}, we introduce our training objective inspired by the Rater Feedback Score (RFS) introduced in~\cite{Visionbased2025}. In this section, we present the detailed definition of the RFS.

In order to calculate the RFS, we define the reference trajectories, provided by dataset, as preferred rater specified trajectories, with a score $\bar{r}$ for each. Suppose the preferred rater specified trajectories are \(W = (w_{1}, w_{2}, ..., w_{H})\). And the predicted waypoints are \(\hat{W} = (\hat{w}_{1},\hat{w}_{2}, ..., \hat{w}_{H})\) as defined in previous paragraph.
With given preferred rater specified trajectories, we therefore define \emph{trust regions}. A trust region is defined for the region within a given lateral and longitudinal threshold of the rater-specified trajectory at a given time $t$ $(t= 3$s, $5$s are used in this work). The size of the trust region depends on the time step of the waypoint (\textbf{time-based thresholds}) and the velocity at the corresponding time (\textbf{speed-based scaling}) in both lateral and longitudinal directions and we finally derive the trust region by following steps:

\paragraph{Time-based thresholds.}  
The raw lateral/longitudinal thresholds $\tilde{\tau}_{\mathrm{lat}},\tilde{\tau}_{\mathrm{lng}}$ (in meters) are defined in Table \ref{tab: Time-based thresholds}.

\begin{table}[ht]
\centering
\begin{tabular}{ccc}
\toprule
Time $t$ & $\tilde{\tau}_{\mathrm{lat, t}}$ & $\tilde{\tau}_{\mathrm{lng, t}}$ \\
\midrule
3 & 1.0 & 4.0 \\
5 & 1.8 & 7.2 \\
\bottomrule
\end{tabular}
\caption{Trust region tolerance values over time}
\label{tab: Time-based thresholds}
\end{table}
\vspace{-2em}

\paragraph{Speed-based scaling.}  
The raw thresholds are scaled by the velocity(m/s) of the preferred rater specified trajectory:
\[
\mathrm{scale}(v)=
\begin{cases}
  0.5, & v<1.4,\\[6pt]
  0.5+0.5\times\frac{v-1.4}{11-1.4}, & 1.4\le v<11,\\[10pt]
  1, & v\ge 11.
\end{cases}
\]
The choice of speed for different scaling comes from the suggestions from~\cite{tefft2010car}. The speed 1.4m/s is around 5kmh (or 3 mph), in which pedestrian crash fatality is relatively low, while 11 m/s is around 40 kmh (or 25mph), in which pedestrian crash fatality reachs around 10\%. So we set up three intervals for the speed of safety, injury, fatality, accordingly.

\paragraph{Final thresholds.} Once we have the raw thresholds and the scale factors, the final thresholds are computed by
\(
\tau_{\mathrm{lat}}(t,v)=\mathrm{scale}(v)\,\tilde{\tau}_{\mathrm{lat,t}}
\)
and 
\(\tau_{\mathrm{lng}}(t,v)=\mathrm{scale}(v)\,\tilde{\tau}_{\mathrm{lng,t}}.\)

If the predicted waypoint is inside the trust region, RFS is assigned to be full score. However, if the predicted waypoint is outside of the trust region, we calculate an exponentially decreasing score for this waypoint. The overall formula is:
\begin{align*}
        \text{RFS}(w_t,\hat{w_t})=
        \begin{cases}
          \bar{r}&,\quad \text{when} \quad\Delta\le 1,\\
          \displaystyle
          \bar{r}\times 0.1^{
            \Delta-1}&, \quad \text{otherwise}.
        \end{cases}
\end{align*}
where $\Delta \doteq\max\left\{\dfrac{\Delta_{\mathrm{lat,t}}}{\tau_{\mathrm{lat,r}}},\dfrac{\Delta_{\mathrm{lng,t}}}{\tau_{\mathrm{lng,t}}}\right\}$ is the maximum distance error among lateral or longitudinal directions, $\Delta_{\mathrm{lat, t}}$ and $\Delta_{\mathrm{lng,t}}$ are lateral and longitudinal distance errors between the predicted waypoint \(\hat{w}_t\) and corresponding waypoint \(w_t\) in preferred rater specified trajecotory at time $t$, and $\bar{r}$ is the full score of preferred rater specified trajectory.

\section{Limitations}
While our proposed framework demonstrates significant improvements in interpretability and trajectory accuracy, we identify two primary limitations that offer avenues for future research.

First, our current architecture processes only the single most recent frame from the multi-view cameras, relying on the explicit numerical state history to capture vehicle kinematics. It does not explicitly encode past video frames. While the state vector provides precise ego-motion data, omitting visual history means the model may miss critical temporal cues regarding the dynamic states of surrounding objects, such as the acceleration of a merging vehicle or the gait of a pedestrian.

Second, although our distilled 8B student model is significantly more efficient than the 78B teacher, achieving real-time control rates remains a challenge due to the autoregressive nature of textual reasoning. Generating a coherent chain-of-thought explanation requires sequentially predicting hundreds of tokens, which introduces a non-negligible latency overhead compared to purely dense-vector architectures. Deploying this system on resource-constrained edge devices or ensuring ultra-low latency for high-speed driving will require further optimization.

{\small
\bibliographystyle{ieee_fullname}
\bibliography{references,references_old_1,references_old_2}
}

\end{document}